\documentclass[11pt,letterpaper]{article}
\usepackage{emnlp2017}
\usepackage{times}
\usepackage{epsfig}
\usepackage{graphicx}
\usepackage{latexsym}
\usepackage{subcaption}
\usepackage{amssymb}
\usepackage{bm}

\usepackage{url}

\emnlpfinalcopy

\def\emnlppaperid{1367}

\title{O\textsc{bj}2T\textsc{ext}: Generating Visually Descriptive Language from \\Object Layouts}

\author{Xuwang Yin \quad Vicente Ordonez \\
  Department of Computer Science, University of Virginia, Charlottesville, VA. \\
  {\tt [xy4cm, vicente]@virginia.edu}
}
\date{}

\begin{document}

\maketitle

\begin{abstract}
   Generating captions for images is a task that has recently received considerable attention. In this work we focus on caption generation for abstract scenes, or object layouts where the only information provided is a set of objects and their locations. We propose OBJ2TEXT, a sequence-to-sequence model that encodes a set of objects and their locations as an input sequence using an LSTM network, and decodes this representation using an LSTM language model. We show that our model, despite encoding object layouts as a sequence, can represent spatial relationships between objects, and generate descriptions that are globally coherent and semantically relevant. We test our approach in a task of object-layout captioning by using only object annotations as inputs. We additionally show that our model, combined with a state-of-the-art object detector, improves an image captioning model from 0.863 to 0.950 (CIDEr score) in the test benchmark of the standard MS-COCO Captioning task.
   
\end{abstract}

\section{Introduction}

\begin{figure}[t!]
  \begin{center}
    \includegraphics[width=0.88\linewidth]{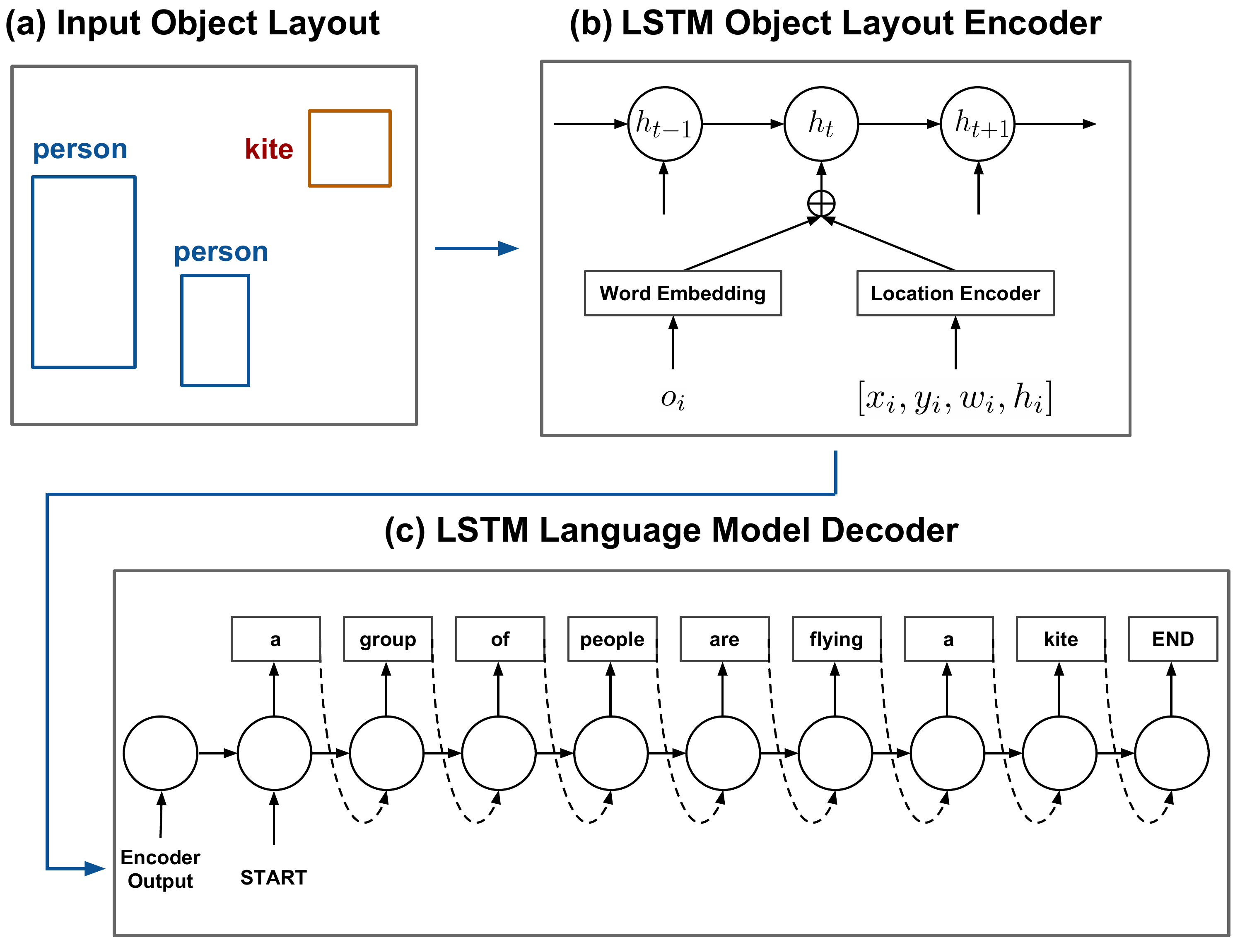}
  \end{center}
  \caption{Overview of our proposed model for generating visually descriptive language from object layouts. The input (a) is an object layout that consists of object categories and their corresponding bounding boxes, the encoder (b) uses a two-stream recurrent neural network to encode the input object layout, and the decoder (c) uses a standard LSTM recurrent neural network to generate text.}
  \label{fig:overview}
\end{figure}

Natural Language generation (NLG) is a long standing goal in natural language processing. There have already been several successes in applications such as financial reporting~\cite{kukich1983design,smadja1990automatically}, or weather forecasts~\cite{konstas2012unsupervised,nlglstm}, however it is still a challenging task for less structured and open domains. Given recent progress in training robust visual recognition models using convolutional neural networks, the task of generating natural language descriptions for arbitrary images has received considerable attention~\cite{showandtell,karpathy,mao2014deep}. In general, generating visually descriptive language can be useful for various tasks such as human-machine communication, accessibility, image retrieval, and search. However this task is still challenging and it depends on developing both a robust visual recognition model, and a reliable language generation model. In this paper, we instead tackle a task of describing object layouts where the categories for the objects in an input scene and their corresponding locations are known. Object layouts are commonly used for story-boarding, sketching, and computer graphics applications. Additionally, using our object layout captioning model on the outputs of an object detector we are also able to improve image captioning models. Object layouts contain rich semantic information, however they also abstract away several other visual cues such as color, texture, and appearance, thus introducing a different set of challenges than those found in traditional image captioning. 

We propose OBJ2TEXT, a sequence-to-sequence model that encodes object layouts using an LSTM network~\cite{hochreiter1997long}, and decodes natural language descriptions using an LSTM-based neural language model\footnote{We build on neuraltalk2 and make our Torch code, and an interactive demo of our model available in the following url: \url{http://vision.cs.virginia.edu/obj2text}}. Natural language generation systems usually consist of two steps: content planning, and surface realization. The first step decides on the content to be included in the generated text, and the second step connects the concepts using structural language properties. In our proposed model, OBJ2TEXT, content planning is performed by the encoder, and surface realization is performed by the decoder. Our model is trained in the standard MS-COCO dataset~\cite{mscoco}, which includes both object annotations for the task of object detection, and textual descriptions for the task of image captioning. While most previous research has been devoted to any one of these two tasks, our paper presents, to our knowledge, the first approach for learning mappings between object annotations and textual descriptions. Using several lesioned versions of the proposed model we explored the effect of object counts and locations in the quality and accuracy of the generated natural language descriptions.

Generating visually descriptive language requires beyond syntax, and semantics; an understanding of the physical word. We also take inspiration from recent work by~\citet{wordordering} where the goal was to reconstruct a sentence from a bag-of-words (BOW) representation using a simple surface-level language model based on an encoder-decoder sequence-to-sequence architecture. In contrast to this previous approach, our model is grounded on visual data, and its corresponding spatial information, so it goes beyond word re-ordering. Also relevant to our work is~\citet{oraclecaptioning} which previously explored the task of oracle image captioning by providing a language generation model with a list of manually defined visual concepts known to be present in the image. In addition, our model is able to leverage both quantity and spatial information as additional cues associated with each object/concept, thus allowing it to learn about verbosity, and spatial relations in a supervised fashion.

In summary, our contributions are as follows:
\begin{itemize}
\item We demonstrate that despite encoding object layouts as a sequence using an LSTM, our model can still effectively capture spatial information for the captioning task. We perform ablation studies to measure the individual impact of object counts, and locations.
\item We show that a model relying only on object annotations as opposed to pixel data, performs competitively in image captioning despite the ambiguity of the setup for this task.
\item We show that more accurate and comprehensive descriptions can be generated on the image captioning task by combining our OBJ2TEXT model using the outputs of a state-of-the-art object detector with a standard image captioning approach.
\end{itemize}


\section{Task}


We evaluate OBJ2TEXT in the task of object layout captioning, and image captioning. In the first task, the input is an object layout that takes the form of a set of object categories and bounding box pairs $\langle\bm{o},\bm{l}\rangle = \{ \langle o_i, l_i \rangle\}$, and the output is natural language. This task resembles the second task of image captioning except that the input is an object layout instead of a standard raster image represented as a pixel array. 
We experiment in the MS-COCO dataset for both tasks. For the first task, object layouts are derived from ground-truth bounding box annotations,  and in the second task object layouts are obtained using the outputs of an object detector over the input image.






\section{Related Work}
Our work is related to previous works that used clipart scenes for visually-grounded tasks including sentence interpretation~\cite{zitnick2013bringing,zitnick2013learning}, and predicting object dynamics~\cite{fouhey2014predicting}. The cited advantage of abstract scene representations such as the ones provided by the clipart scenes dataset proposed in~\cite{zitnick2013bringing} is their ability to separate the complexity of pattern recognition from semantic visual representation. Abstract scene representations also maintain common-sense knowledge about the world. The works of~\citet{vedantam2015learning,eysenbach2016mistaken} proposed methods to learn common-sense knowledge from clipart scenes, while the method of~\citet{yatskar2016stating}, similar to our work, leverages object annotations for natural images. Understanding abstract scenes has demonstrated to be a useful capability for both language and vision tasks and our work is another step in this direction.

Our work is also related to other language generation tasks such as image and video captioning~\cite{farhadi10,im2text,mason2014nonparametric,bigdataIJCV2015,Xu2015ShowAA,Donahue2015LongtermRC,mao2014deep,Fang2015FromCT}. This problem is interesting because it combines two challenging but perhaps complementary tasks: visual recognition, and generating coherent language. Fueled by recent advances in training deep neural networks~\cite{alexnet} and the availability of large annotated datasets with images and captions such as the MS-COCO dataset~\cite{mscoco}, recent methods on this task perform end-to-end learning from pixels to text. Most recent approaches use a variation of an encoder-decoder model where a convolutional neural network (CNN) extracts visual features from the input image (encoder), and passes its outputs to a recurrent neural network (RNN) that generates a caption as a sequence of words (decoder)~\cite{karpathy,showandtell}. However, the MS-COCO dataset, containing object annotations, is also a popular benchmark in computer vision for the task of object detection, where the objective is to go from pixels to a collection of object locations. In this paper, we instead frame our problem as going from a collection of object categories and locations (object layouts) to image captions. This requires proposing a novel encoding approach to encode these object layouts instead of pixels, and allows for analyzing the image captioning task from a different perspective. Several other recent works use a similar sequence-to-sequence approach to generate text from source code input~\cite{lukecode}, or to translate text from one language to another~\cite{bahdanau2014neural}.

There have also been a few previous works explicitly analyzing the role of spatial and geometric relations between objects for vision and language related tasks. The work of~\citet{elliott2013image} manually defined a dictionary of object-object relations based on geometric cues. The work of~\citet{ramisa2015combining} is focused on predicting preposition given two entities and their locations in an image. 
Previous works of~\citet{flickr30kentities} and~\citet{groundreconstruction} showed that switching from classification-based CNN network to detection-based Fast RCNN network improves performance for phrase localization. The work of~\citet{hu2016natural} showed that encoding image regions with spatial information is crucial for natural language object retrieval as the task explicitly asks for locations of target objects. 
Unlike these previous efforts, our model is trained end-to-end for the language generation task, and takes as input a holistic view of the scene layout, potentially learning higher order relations between objects.

\section{Model}
In this section we describe our base OBJ2TEXT model for encoding object layouts to produce text (section~\ref{sec:scene_layout_decoder}), as well as two further variations to use our model to generate captions for real images: OBJ2TEXT-YOLO which uses the YOLO object detector~\cite{yolo2} to generate layouts of object locations from real images (section~\ref{sec:obj2text-yolo}), and OBJ2TEXT-YOLO + CNN-RNN which further combines the previous model with an encoder-decoder image captioning which uses a convolutional neural network to encode the image (section~\ref{sec:obj2text-yolo-showandtell}).

\subsection{OBJ2TEXT} 
\label{sec:scene_layout_decoder} 
OBJ2TEXT is a sequence-to-sequence model that encodes an input object layout as a sequence, and decodes a textual description by predicting the next word at each time step. Given a training data set comprising $N$ observations $\left \{  \langle \bm{o}^{(n)},\bm{l}^{(n)} \rangle \right \}$, where $\langle \bm{o}^{(n)},\bm{l}^{(n)} \rangle$ is a pair of sequences of input category and location vectors, together with a corresponding set of target captions $\left \{\bm{s}^{(n)} \right \}$, the encoder and decoder are trained jointly by minimizing a loss function over the training set using stochastic gradient descent:
\begin{equation}
    {W}^{\ast} = \arg\min_{{W}} \sum_{n=1}^N \mathcal{L}(\langle\bm{o}^{(n)},\bm{l}^{(n)}\rangle, \bm{s}^{(n)}),
\end{equation}
in which $W={{W}_1 \choose {W}_2}$ is the group of encoder parameters ${W}_1$ and decoder parameters ${W}_2$. The loss function is a negative log likelihood function of the generated description given the encoded object layout
\begin{equation}
    \mathcal{L}(\langle\bm{o}^{(n)},\bm{l}^{(n)}\rangle, \bm{s}^{(n)}) = -\log p(\bm{s}^{n} | h_L^n,W_2),
\end{equation}
where $h_L^n$ is computed using the LSTM-based encoder (eqs.~\ref{eq:xt}, and~\ref{eq:ht1}) from the object layout inputs $\langle \bm{o}^{(n)},\bm{l}^{(n)} \rangle$, and $p(\bm{s}^{n} | h_L^n,W_2)$ is computed using the LSTM-based decoder (eqs.~\ref{eq:ps},~\ref{eq:pst} and~\ref{eq:ht}).

At inference time we encode an input layout $\langle \bm{o},\bm{l} \rangle$ into its representation $h_L$, and sample a sentence word by word based on  $p(s_t|h_L,\bm{s}_{<t})$ as computed by the decoder in  time-step $t$. Finding the optimal sentence ${\bm{s}}^{\ast} = \arg\max_{{\bm{s}}} p(\bm{s}|h_L)$ requires the evaluation of an exponential number of sentences as in each time-step we have $K$ number of choices for a word vocabulary of size $K$. As a common practice for an approximate solution, we follow~\cite{showandtell} and use beam search to limit the choices for words at each time-step by only using the ones with the highest probabilities.

\vspace{0.2em}
\noindent{\bf Encoder}: The encoder at each time-step $t$ takes as input a pair $\langle o_t, l_t \rangle$, where $o_t$ is the object category encoded as a one-hot vector of size $V$, and $l_t = [B^x_t, B^y_t, B^w_t, B^h_t]$ is the location configuration vector that contains left-most position, top-most position, and the width and height of the bounding box corresponding to object $o_t$, all normalized in the range [0,1] with respect to input image dimensions. 
$o_t$ and $l_t$ are mapped to vectors with the same size $k$ and added to form the input $x_t$ to one time-step of the LSTM-based encoder as follows:
\begin{equation}
x_t = W_o o_t + (W_l l_t +b_l) \textit{,  }\quad x_t \in \mathbb{R}^k,
\label{eq:xt}
\end{equation}
in which $W_o \in \mathbb{R}^{k\times V} $ is a categorical embedding matrix (the word encoder), and $W_l \in \mathbb{R}^{k\times 4}$ and bias $b_l \in \mathbb{R}^k$ are parameters of a linear transformation unit (the object location encoder).

Setting initial value of cell state vector $c_0^{e} = 0$ and hidden state vector $h_0^{e} = 0$, the LSTM-based encoder takes the sequence of input $(x_1,...,x_{T_1} )$ and generates a sequence of hidden state vectors $( h_1^{e},...,h_{T_1}^{e})$ using the following step function (we omit cell state variables and internal transition gates for simplicity as we use a standard LSTM cell definition):
\begin{equation}
    h_t^{e} = \textnormal{LSTM}(h_{t-1}^{e},x_t; {W}_1).
    \label{eq:ht1}
\end{equation}
We use the last hidden state vector  $h_L = h^{e}_{T_1}$ as the encoded representation of the input layout $\langle \bm{o}_t, \bm{l}_t \rangle$ to generate the corresponding description $\bm{s}$. 

\vspace{0.2em}
\noindent{\bf Decoder}: The decoder takes the encoded layout $h_L$ as input and generates a sequence of multinomial distributions over a vocabulary of words using an LSTM neural language model. The joint probability distribution of generated sentence $\bm{s}=(s_1,...,s_{T_2})$ is factorized into products of conditional probabilities:
\begin{equation}
    p(\bm{s}|h_L) = \prod_{t=1}^{T_2} p(s_t|h_L,\bm{s}_{<t}),
    \label{eq:ps}
\end{equation}
where each factor is computed using a softmax function over the hidden states of the decoder LSTM as follows:
\begin{equation}
    p(s_t|h_L,\bm{s}_{<t}) = \textnormal{softmax}(W_h h_{t-1}^{d} + b_h),
    \label{eq:pst}
\end{equation}
\begin{equation}
    h_t^{d} = \textnormal{LSTM}(h^{d}_{t-1},W_s s_{t}; {W}_2),
    \label{eq:ht}
\end{equation}
where $W_s$ is the categorical embedding matrix for the one-hot encoded caption sequence of symbols. By setting $h_{-1}^{d} = 0$ and $c_{-1}^{d}=0$ for the initial hidden state and cell state, the layout representation is encoded into the decoder network at the $0$ time step as a regular input:
\begin{equation}
    h_0^{d} = \textnormal{LSTM} (h_{-1}^{d},h_L; {W}_2).
    \label{eq:h0}
\end{equation}
We use beam search to sample from the LSTM as is routinely performed in previous literature in order to generate text.

\subsection{OBJ2TEXT-YOLO}
\label{sec:obj2text-yolo}
For the task of image captioning we propose OBJ2TEXT-YOLO. This model takes an image as input, extracts an object layout (object categories and locations) with a state-of-the-art object detection model YOLO~\cite{yolo2}, and uses OBJ2TEXT as described in section~\ref{sec:scene_layout_decoder} to generate a natural language description of the input layout and hence, the input image. The model is trained using the standard back-propagation algorithm, but the error is not back-propagated to the object detection module.

\subsection{OBJ2TEXT-YOLO + CNN-RNN}
\label{sec:obj2text-yolo-showandtell}
\begin{figure}[t!]
  \begin{center}
    \includegraphics[width=0.99\linewidth]{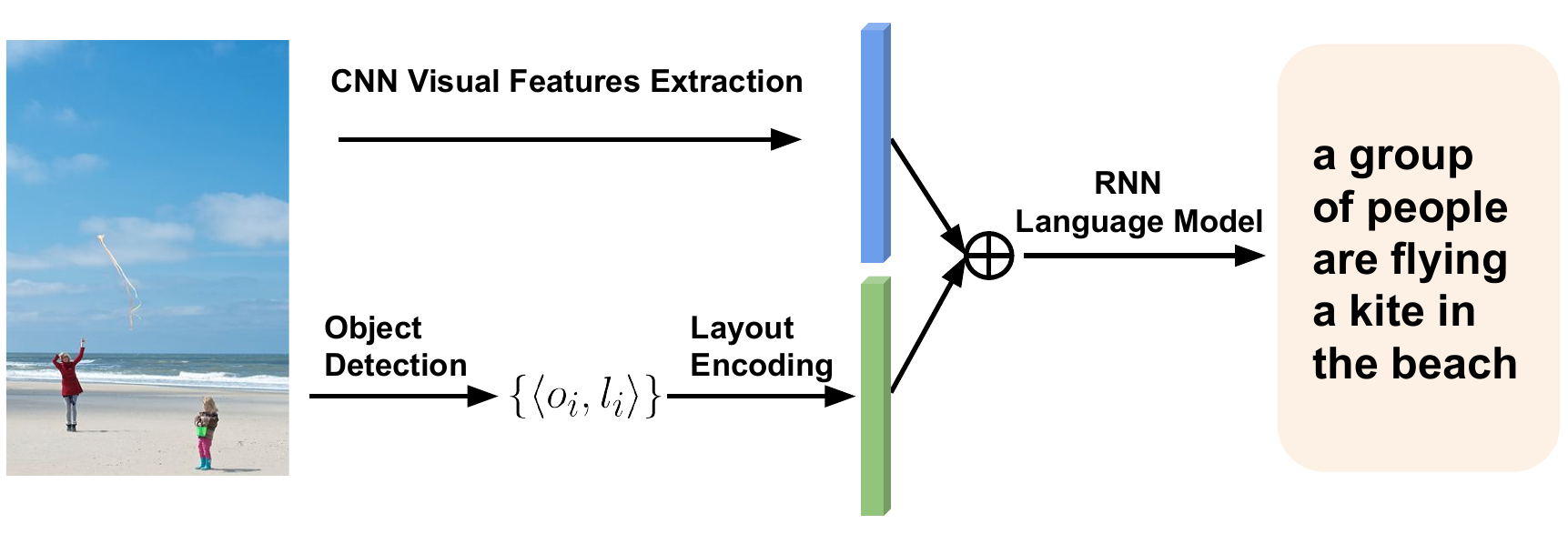}
  \end{center}
  \caption{Image Captioning by joint learning of visual features and object layout encoding.}
  \label{fig:multisource_captioning}
\end{figure}

For the image captioning task we experiment with a combined model (see Figure~\ref{fig:multisource_captioning}) where we take an image as input, and then use two separate computation branches to extract visual feature information and object layout information. These two streams of information are then passed to an LSTM neural language model to generate a description. Visual features are extracted using the VGG-16~\cite{vgg} convolutional neural network  pre-trained on the ImageNet classification task~\cite{russakovsky2014imagenet}. Object layouts are extracted using the YOLO object detection system and its output object locations are encoded using our proposed OBJ2TEXT encoder. These two streams of information are encoded into vectors of the same size and their sum is input to the language model to generate a textual description. The model is trained using the standard back-propagation algorithm where the error is back-propagated to both branches but not the object detection module. The weights of the image CNN model are fine-tuned only after the layout encoding branch is well trained but no significant overall performance improvements were observed.

\section{Experimental Setup}
\label{sec:setup}

We evaluate the proposed models on the MS-COCO~\cite{mscoco} dataset which is a popular image captioning benchmark that also contains object extent annotations. In the object layout captioning task the model uses the ground-truth object extents as input object layouts, while in the image captioning task the model takes raw images as input. The qualities of generated descriptions are evaluated using both human evaluations and automatic metrics. 
We train and validate our models based on the commonly adopted split regime (113,287 training images, 5000 validation and 5000 test images) used in~\cite{neuraltalk2}, and also test our model in the MS-COCO official test benchmark. 

We implement our models based on the open source image captioning system Neuraltalk2~\cite{neuraltalk2}. Other configurations including data preprocessing and training hyper-parameters also follow Neuraltalk2. We trained our models using a GTX1080 GPU with 8GB of memory for 400k iterations using a batch size of 16 and an Adam optimizer with alpha of 0.8, beta of 0.999 and epsilon of 1e-08. 
Descriptions of the CNN-RNN approach are generated using the publicly available code and model checkpoint provided by Neuraltalk2~\cite{neuraltalk2}. Captions for online test set evaluations are generated using beam search of size 2, but score histories on split validation set are based on captions generated without beam search (i.e. max sampling at each time-step).


\noindent{\bf  Ablation on Object Locations and Counts:}.
We setup an experiment where we remove the input locations from the OBJ2TEXT encoder to study the effects on the generated captions, and confirm whether the model is actually using spatial information during surface realization. In this restricted version of our model the LSTM encoder at each time step only takes the object category embedding vector as input. The OBJ2TEXT model additionally encodes different instances of the same object category in different time steps, potentially encoding in some of its hidden states information about how many objects of a particular class are in the image. For example, in the object annotation presented in the input in Figure~\ref{fig:overview}, there are two instances of ``person''. We perform an additional experiment where our model does not have access neither to object locations, nor the number of object instances by providing only a set of object categories. Note that in this set of experiments the object layouts are given as inputs, thus we assume full access to ground-truth object annotations, even in the test split. In the experimental results section we use the ``-GT''  postfix to indicate that input object layouts are obtained from ground-truth object annotations provided by the MS-COCO dataset. 

\noindent{\bf Image Captioning Experiment:}
 In this experiment we assess whether the image captioning model OBJ2TEXT-YOLO that only relies on object categories and locations could give comparable performance with a CNN-RNN model based on Neuraltalk2~\cite{neuraltalk2} that has full access to visual image features. We also explore how much does a combined OBJ2TEXT-YOLO + CNN-RNN model could improve over a CNN-RNN model by fusing object counts and location information that is not explicitly encoded in a traditional CNN-RNN approach. 

\noindent{\bf Human Evaluation Protocol}.
We use a two-alternative forced-choice evaluation (2AFC) approach to compare two methods that generate captions. For this, we setup a task on Amazon Mechanical Turk where users are presented with an image and two alternative captions, and they have to choose the caption that best describes the image. Users are not prompted to use any single criteria but rather a holistic assessment of the captions, including their semantics, syntax, and the degree to which they describe the image content. In our experiment we randomly sample 500 captions generated by various models for MS COCO online test set images, and use three users per image to obtain annotations. 
Note that three users choosing randomly between two options have a chance of 25\% to select the same caption for a given image. In our experiments comparing method $A$ vs method $B$, we report the percentage of times $A$ was picked over $B$ (Choice-all), the percentage of times all users selected the same method, either $A$ or $B$, (Agreement), and the percentage of times $A$ was picked over $B$ only for these cases where all users agreed (Choice-agreement).

\section{Results}
\label{sec:results}

\begin{figure*}[th!]
    \centering
    \begin{subfigure}[t]{0.5\textwidth}
        \centering
        \includegraphics[width=\linewidth]{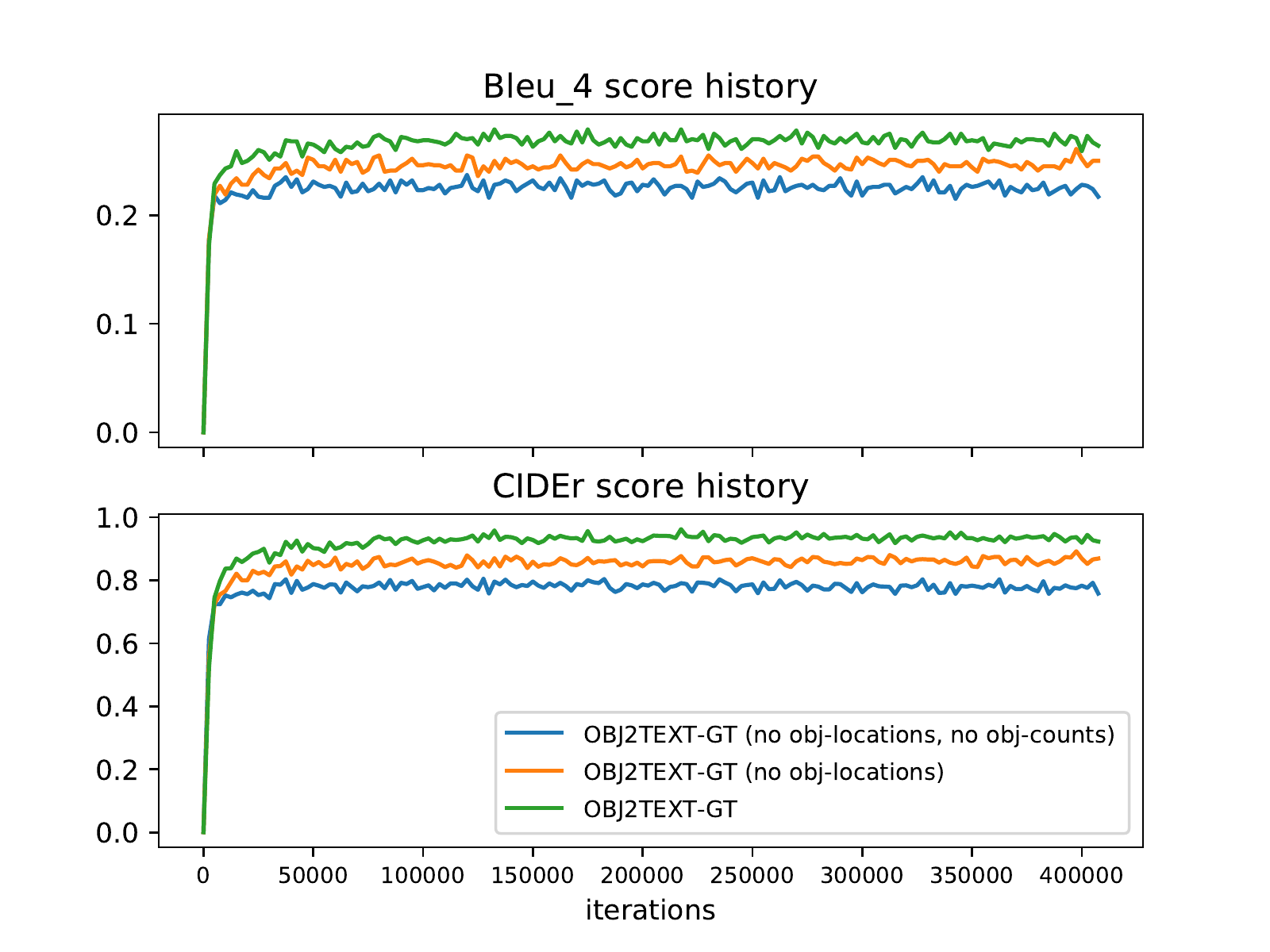}
        \caption{Score histories of lesioned versions of the proposed model for the task of object layout captioning.}
        \label{fig:location_compare}
    \end{subfigure}%
    ~ 
    \begin{subfigure}[t]{0.5\textwidth}
        \centering
        \includegraphics[width=\linewidth]{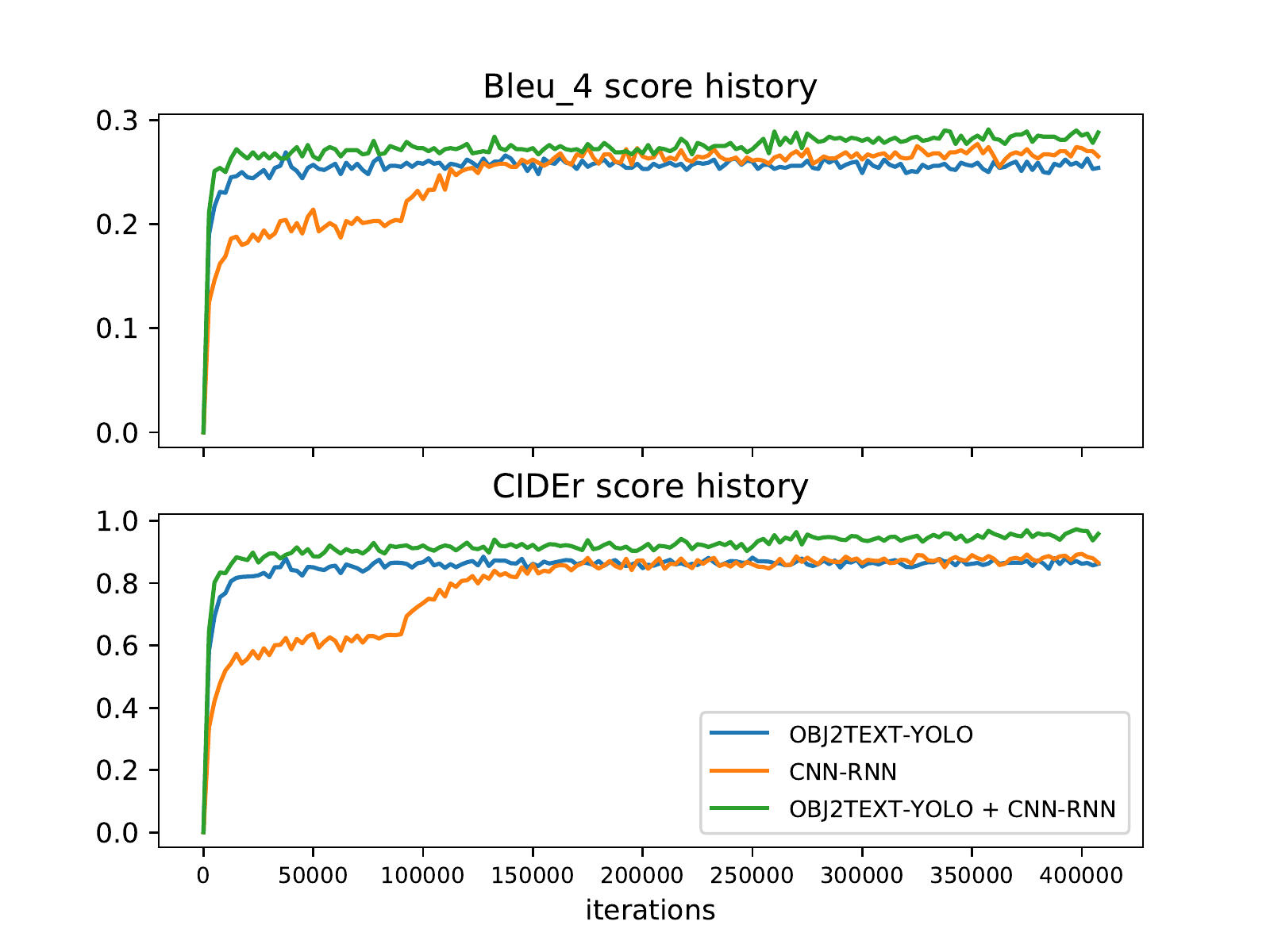}
        \caption{Score histories of image captioning models.
         Performance boosts of CNN-RNN and combined model around iteration 100K and 250K are due to fine-tuning of the image CNN model. 
         }
        \label{fig:yolo_combine_compare}
    \end{subfigure}
    \caption{Score histories of various models on the MS COCO split validation set.}
    \label{fig:history}
\end{figure*}

\begin{table*}[t]
\centering
\begin{tabular}{lrrrr}
\hline
Method                                 & Bleu\_4        & CIDEr          & METEOR         & ROUGE-L        \\ \hline
OBJ2TEXT-GT (no obj-locations, counts) & 0.21           & 0.759          & 0.215          & 0.464          \\
OBJ2TEXT-GT (no obj-locations)         & 0.233          & 0.837          & 0.222          & 0.482          \\
OBJ2TEXT-GT                            & \textbf{0.253} & \textbf{0.922} & \textbf{0.238} & \textbf{0.507} \\ \hline
\end{tabular}
  \caption{Performance of lesioned versions of the proposed model on the MS COCO split test set. 
  }
  \label{table:test_scores}
\end{table*}



  
  
  
 
  
\begin{table*}
  \centering
  \begin{tabular}{lrrr}
Alternatives                                   & Choice-all & Choice-agreement & Agreement \\ \hline
OBJ2TEXT-GT vs. OBJ2TEXT-GT (no obj-locations) & 54.1\%     & 62.1\%         & 40.6\%    \\ \hline
OBJ2TEXT-YOLO vs. CNN+RNN                      & 45.6\%     & 40.6\%         & 54.7\%      \\
OBJ2TEXT-YOLO + CNN-RNN vs. CNN-RNN              & 58.1\%     & 65.3\%         & 49.5\%    \\ \hline
 OBJ2TEXT-GT vs. HUMAN                          & 23.6\%     & 9.9\%          & 58.8\%    \\ \hline
  \end{tabular}

  \caption{Human evaluation results using two-alternative forced choice evaluation. Choice-all is percentage the first alternative was chosen. Choice-agreement is percentage the first alternative was chosen only when all annotators agreed. Agreement is percentage where all annotators agreed (random is 25\%). }
  \label{table:human_evaluation}
\end{table*}

\begin{table*}[t]

\begin{tabular}{llllllll}
\hline
MS COCO Test Set Performance & CIDEr          & ROUGE-L        & METEOR         & B-4            & B-3            & B-2            & B-1            \\ \hline
\textbf{5-Refs}           &                &                &                &                &                &                &                \\ \hline
OBJ2TEXT-YOLO                & 0.830          & 0.497          & 0.228          & 0.262          & 0.361          & 0.500          & 0.681          \\
CNN-RNN                      & 0.857          & 0.514          & 0.237          & 0.283          & 0.387          & 0.529          & 0.705          \\
OBJ2TEXT-YOLO + CNN-RNN        & \textbf{0.932} & \textbf{0.528} & \textbf{0.250} & \textbf{0.300} & \textbf{0.404} & \textbf{0.546} & \textbf{0.719} \\ \hline
\textbf{40-Refs}          &                &                &                &                &                &                &                \\ \hline
OBJ2TEXT-YOLO                & 0.853          & 0.636          & 0.305          & 0.508          & 0.624          & 0.746          & 0.858          \\
CNN-RNN                      & 0.863          & 0.654          & 0.318          & 0.540          & 0.656          & 0.775          & 0.877          \\
OBJ2TEXT-YOLO + CNN-RNN        & \textbf{0.950} & \textbf{0.671} & \textbf{0.334} & \textbf{0.569} & \textbf{0.686} & \textbf{0.802} & \textbf{0.896} \\ \hline
\end{tabular}
  \caption{The 5-Refs and 40-Refs performances of OBJ2TEXT-YOLO, CNN-RNN and the combined approach on the MS COCO online test set. The 5-Refs performance is measured using 5 ground-truth reference captions, while 40-Refs performance is measured using 40 ground-truth reference captions.
  }
  \label{table:online_test_scores}
\end{table*}

\noindent{\bf Impact of Object Locations and Counts:}
Figure~\ref{fig:location_compare} shows the CIDEr~\cite{vedantam2015cider}, and BLEU-4~\cite{papineni2002bleu} score history on our validation set during 400k iterations of training of OBJ2TEXT, as well as a version of our model that does not use object locations, and a version of our model that does not use neither object locations nor object counts. These results show that our model is effectively using both object locations and counts to generate better captions, and absence of any one of these two cues affects performance. Table~\ref{table:test_scores} confirms these results on the test split after a full round of training. 

\begin{figure*}[t!]
  \begin{center}
    \includegraphics[width=0.9\linewidth]{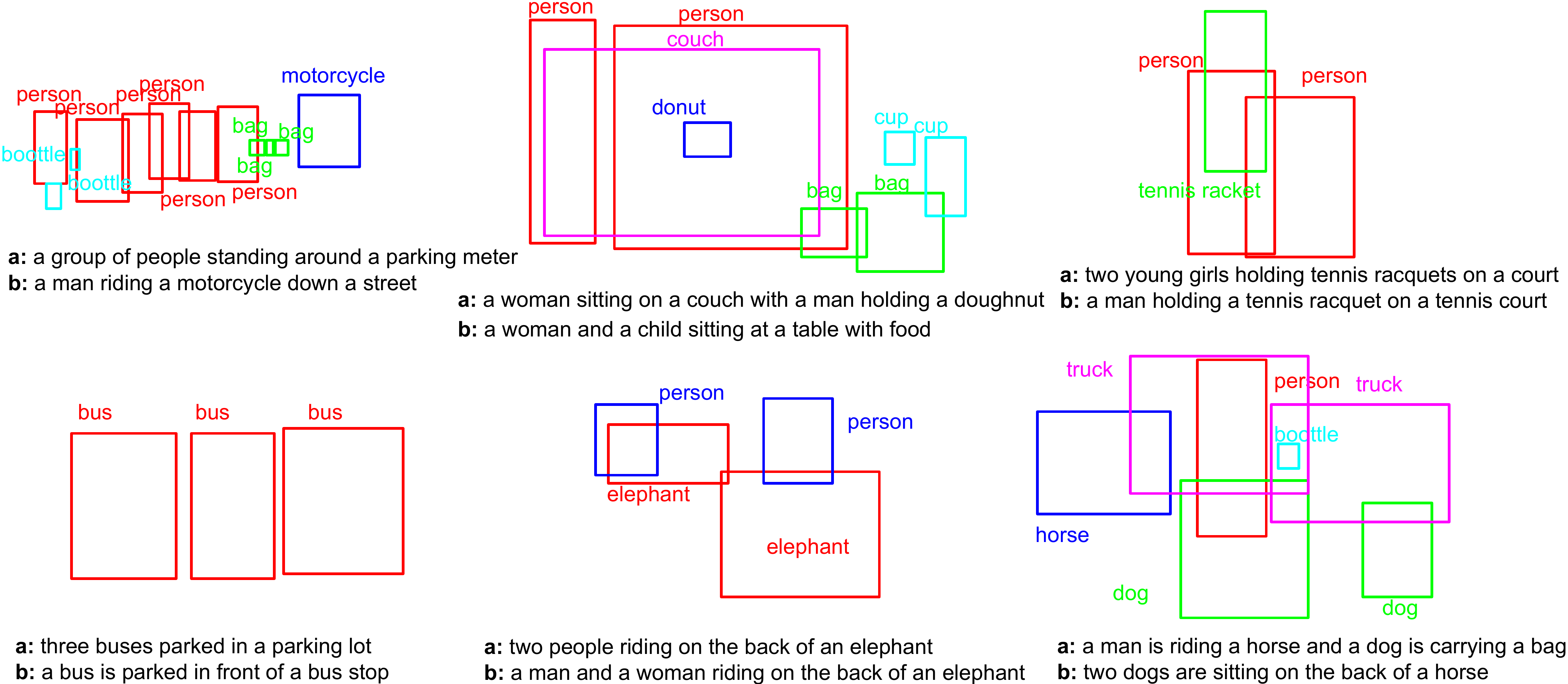}
  \end{center}
  \caption{Qualitative examples comparing generated captions of (\textbf{a}) OBJ2TEXT-GT, and (\textbf{b}) OBJ2TEXT-GT (no obj-locations).}
  \label{fig:samples_vs_objname}
\end{figure*}

Furthermore, human evaluation results in the first row of Table~\ref{table:human_evaluation}
 show that the OBJ2TEXT model with access to object locations is preferred by users, especially in cases where all evaluators agreed on their choice (62\% over the baseline that does not have access to locations). In Figure~\ref{fig:samples_vs_objname} we additionally present qualitative examples showing predictions side-by-side between OBJ2TEXT-GT and OBJ2TEXT-GT (no obj-locations).  These results indicate that 1) perhaps not surprisingly, object counts is useful for generating better quality descriptions, and 2) object location information when properly encoded, is an important cue for generating more accurate descriptions.  We additionally implemented a nearest neighbor baseline by representing the objects in the input layout using an orderless bag-of-words representation of object counts and the CIDEr score on the test split was only 0.387.

On top of OBJ2TEXT we additionally experimented with the global attention model proposed in~\cite{LuongPM15} so that a weighted combination of the encoder hidden states are forwarded to the decoding neural language model, however we did not notice any overall gains in terms of accuracy from this formulation. We observed that this model provided gains only for larger input sequences where it is more likely that the LSTM network forgets its past history~\cite{bahdanau2014neural}. However in MS-COCO the average number of objects in each image is rather modest, so the last hidden state can capture well the overall nuances of the visual input.

\noindent{\bf Object Layout Encoding for Image Captioning:}
Figure~\ref{fig:yolo_combine_compare} shows the CIDEr, and BLEU-4 score history on the validation set during 400k iterations of training of OBJ2TEXT-YOLO, CNN-RNN, and their combination. These results show that OBJ2TEXT-YOLO performs surprisingly close to CNN-RNN, and the model resulting from combining the two, clearly outperforms each method alone. Table~\ref{table:online_test_scores} shows MS-COCO evaluation results on the test set using their online benchmark service, and confirms results obtained in the validation split, where CNN-RNN seems to have only a slight edge over OBJ2TEXT-YOLO which lacks access to pixel data after the object detection stage. Human evaluation results in Table~\ref{table:human_evaluation} rows 2, and 3, further confirm these findings. These results show that meaningful descriptions could be generated solely based on object categories and locations information, even without access to color and texture input.

The combined model performs better than the two models, improving the CIDEr score of the basic CNN-RNN model from 0.863 to 0.950, and human evaluation results show that the combined model is preferred over the basic CNN-RNN model for 65.3\% of the images for which all evaluators were in agreement about the selected method.
These results show that explicitly encoded object counts and location information, which is often overlooked in traditional image captioning approaches, could boost the performance of existing models. Intuitively, object layout and visual features are complementary: neural network models for visual feature extraction are trained on a classification task where object-level information such as number of instances and locations are ignored in the objective. Object layouts on the other hand, contain categories and their bounding-boxes but don't have access to rich image features such as image background, object attributes and objects with categories not present in the object detection vocabulary. 

\begin{figure*}[p]
  \begin{center}
    \includegraphics[width=\linewidth]{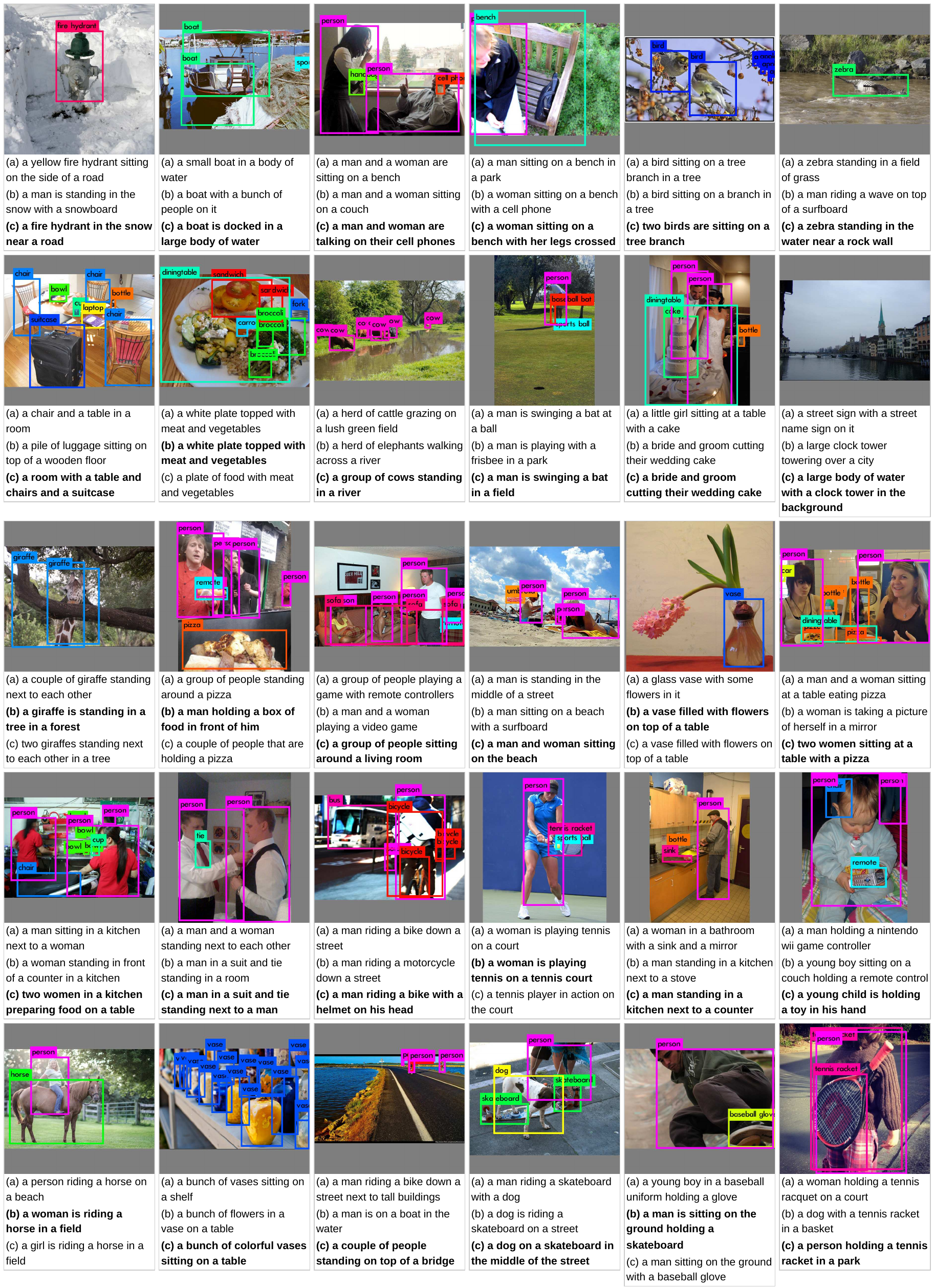}
  \end{center}
  \caption{Qualitative examples comparing the generated captions of  (a) OBJ2TEXT-YOLO, (b) CNN-RNN  and (c) OBJ2TEXT-YOLO + CNN-RNN. Images are selected from the 500 human evaluation images and annotated with YOLO object detection results. Captions preferred by human evaluators with agreement are highlighted in bold text.}
  \label{fig:3way}
\end{figure*}

Figure~\ref{fig:3way} provides a three-way comparison of captions generated by the three image captioning models, with preferred captions by human evaluators annotated in bold text. Analysis on actual outputs gives us insights into the benefits of combing object layout information and visual features obtained using a CNN. Our OBJ2TEXT-YOLO model makes many mistakes because of lack of image context information since it only has access to object layout, while CNN-RNN makes many mistakes because the visual recognition model is imperfect at predicting the correct content. The combined model is usually able to generate more accurate and comprehensive descriptions.

In this work we only explored encoding spatial information with object labels, but object labels could be readily augmented with rich semantic features that are more detailed descriptions of objects or image regions. For example, the work of~\citet{you2016image} and~\citet{yao2016boosting} showed that visual features trained with semantic concepts (text entities mentioned in captions) instead of object labels is useful for image captioning, although they didn't consider encoding semantic concepts with spatial information. In case of object annotations the MS-COCO dataset only provides object labels and bounding-boxes, but there are other datasets such as Flick30K Entities~\cite{flickr30kentities}, and the Visual Genome dataset~\cite{krishnavisualgenome} that provide richer region-to-phrase correspondence annotations. In addition, the fusion of object counts and spatial information with CNN visual features could in principle benefit other vision and language tasks such as visual question answering.  We leave these possible extensions as future work.

\section{Conclusion}
\label{sec:length}

We introduced OBJ2TEXT, a sequence-to-sequence model to generate visual descriptions for object layouts where only categories and locations are specified. Our proposed model shows that an orderless visual input representation of concepts is not enough to produce good descriptions, but object extents, locations, and object counts, all contribute to generate more accurate image descriptions. Crucially we show that our encoding mechanism is able to capture useful spatial information using an LSTM network to produce image descriptions, even when the input is provided as a sequence rather than as an explicit 2D representation of objects. Additionally, using our proposed OBJ2TEXT model in combination with an existing image captioning model and a robust object detector we showed improved results in the task of image captioning.

\vspace{-0.05in}
\section*{Acknowledgments}
\vspace{-0.05in}
This work was supported in part by an NVIDIA Hardware Grant. We are also thankful for the feedback from Mark Yatskar and anonymous reviewers of this paper.

\bibliography{emnlp2017,references_vicente}
\bibliographystyle{emnlp_natbib}

\end{document}